# Exceptional Subclasses in Qualitative Probability

Sek-Wah Tan
Cognitive Systems Lab
University of California
Los Angeles, CA 90024
< tan@cs.ucla.edu >

## Abstract

System $Z^+$ [Goldszmidt and Pearl, 1991, Goldszmidt, 1992] is a formalism for reasoning with normality defaults of the form "typically if $\varphi$ then $\psi$ (with strength $\delta$)" where $\delta$ is a positive integer. The system has a critical shortcoming in that it does not sanction inheritance across exceptional subclasses. In this paper we propose an extension to System $Z^+$ that rectifies this shortcoming by extracting additional conditions between worlds from the defaults database. We show that the additional constraints do not change the notion of the consistency of a database. We also make comparisons with competing default reasoning systems.

## 1 Introduction

Goldszmidt's system $Z^+$ [Goldszmidt and Pearl, 1991, Goldszmidt, 1992] is a formalism for reasoning with normality defaults of the form "typically if $\varphi$ then $\psi$ (with strength $\delta$)" where $\delta$ is a non-negative integer. The system is attractive because it has been shown to be semi-tractable [Goldszmidt and Pearl, 1992] in that it is tractable for every sublanguage in which propositional satisfiability is tractable. There is however a critical shortcoming with $Z^+$ when compared to competing accounts like conditional entailment [Geffner, 1989], system $Z^*$ [Goldszmidt et al., 1990, Goldszmidt, 1992] (maximum entropy) and $CO^*$ [Boutilier, 1992] (when augmented with Brewka's preferred subtheories [Brewka, 1989]). $Z^+$ does not sanction inheritance across exceptional subclasses. For example, consider a defaults database $\Delta$ containing the normality defaults "typically birds fly" $b \to f$, "penguins typically do not fly" $p \to \neg f$, "penguins typically are birds" $p \to b$ and "birds typically have wings" $b \to w$. $Z^+$ will conclude that penguins do not fly, and once penguins have been determined to be exceptional birds with respect to flying, penguins will be considered as being exceptional birds with respect to all properties. In particular $Z^+$ will not be able to conclude that penguins have wings, a conclusion that we would like to draw from $\Delta$.

In this paper we propose a refinement to system $Z^+$ to rectify this shortcoming. $Z^+$ assigns to each world an integral *belief rank* that represents the believability of the world and interprets each normality default as a constraint on the belief ranks. In addition to the $Z^+$ constraints, we impose new conditions called *cp-conditions* that ensures the following (whenever possible): every world that is inconsistent with a normality default to be ranked higher (less believable) than a world that is consistent with the default when the worlds agree *ceteris paribus* on all the other normality defaults in the database. We will consider the penguin example to illustrate how penguins, despite being exceptional birds with respect to flying, can inherit other properties of typical birds. We will also show that *cp-consistency* in this stronger system turns out to be no different from the original idea of consistency in system $Z^+$. This implies that the procedure for testing consistency presented in [Goldszmidt, 1992] may also be used to test for cp-consistency. Thus the complexity of deciding cp-consistency is no worse than that of propositional satisfiability and is tractable for useful sublanguages like Horn clauses.

In section 2 we describe system $Z^+$ and the constraints it imposes on the belief rankings. In section 3 we introduce the notions of *cp-conditions*, *cp-admissibility* and *cp-consistency*. We will prove the equivalence of cp-consistency and consistency in system $Z^+$ and also define the $\bar{\kappa}$ belief ranking which embodies the assumption of maximal ignorance. A number of related systems are compared in section 4 before we conclude with a summary of the paper.

## 2 System $Z^+$

System $Z^+$ considers normality defaults of the form "typically, if $\varphi$ then $\psi$ (with strength $\delta$)" where $\delta$ is a non-negative integer and written $\varphi \xrightarrow{\delta} \psi$. A normality default with an unspecified $\delta$ will be assumed to have a strength of 0. Normality defaults are interpreted as constraints on the believability (or belief rank) of worlds. A belief rank of 0 indicates that the world is believable and higher ranks represent higher degrees of abnormality (or surprise), that is decreasing believability.

Given a normality default $d = \varphi \xrightarrow{\delta} \psi$ and a world $\omega$ we say that $\omega$ *verifies* $d$ if $\omega \models \varphi \wedge \psi$. $\omega$ *falsifies* $d$ if $\omega \models \varphi \wedge \neg\psi$ and $\omega$ *satisfies* $d$ if $\omega \models \varphi \supset \psi$. We also say that the worlds $\omega$ and $\nu$ *agree* (ceteris paribum) on $d$ if

1. $\omega$ verifies $d$ if and only if $\nu$ verifies $d$, and
2. $\omega$ falsifies $d$ if and only if $\nu$ falsifies $d$.

The default $d$ is *tolerated* by a set of normality defaults $S$ if the wff $\varphi \wedge \psi \bigwedge_i \varphi_i \supset \psi_i$ is satisfiable (where $i$ ranges over all rules in $S$).

**Definition 1 (Belief Ranking)** *A belief ranking $\kappa$ is a non-negative integer-valued function on the set of worlds $\Omega$ such that $\kappa(\omega) = 0$ for some world $\omega \in \Omega$. Given wffs $\varphi$ and $\psi$, the belief ranking can be extended as follows:*

$$\kappa(\varphi) = \begin{cases} \min_{\omega \models \varphi} \kappa(\omega) & \text{if } \varphi \text{ is satisfiable} \\ \infty & \text{otherwise.} \end{cases}$$

$$\kappa(\psi|\varphi) = \begin{cases} \kappa(\varphi \wedge \psi) - \kappa(\varphi) & \text{if } \kappa(\varphi) \neq \infty \\ \infty & \text{otherwise.} \end{cases}$$

A belief ranking $\kappa$ may be considered to be an order-of-magnitude approximation of a probability function $P$ by writing it as a polynomial of some small number $\epsilon$ and considering only the most significant term in the polynomial,

$$P(\omega) \approx C\epsilon^{\kappa(\omega)}.$$

Intuitively, if $\kappa$ reflects our beliefs about worlds accurately then $\kappa(\psi|\varphi)$ represents the degree of abnormality or surprise associated with discovering $\psi$, given that we already know $\varphi$, and $\kappa(\psi) < \kappa(\varphi)$ indicates that $\psi$ is more believable than $\varphi$.

**Definition 2 (Admissibility)** *Let $\Delta$ be a set of normality defaults. A belief ranking is **admissible** with respect to $\Delta$ if and only if*

$$\kappa(\varphi \wedge \neg\psi) > \kappa(\varphi \wedge \psi) + \delta$$

*for every normality default $\varphi \xrightarrow{\delta} \psi \in \Delta$. $\Delta$ is said to be **consistent** if it has an admissible belief ranking.*

Thus a normality default $\varphi \xrightarrow{\delta} \psi$ imposes the constraint that given $\varphi$ it would be surprising by at least $\delta$ degrees to find $\neg\psi$. This reflects the usual interpretation of defaults where $\psi$ holds in all the minimal (preferred or most believed) models for $\varphi$.

This interpretation of defaults turns out to be too weak. The admissibility constraints imposed by a normality default $\varphi \xrightarrow{\delta} \psi$ constrains only the most preferred $\varphi$-worlds, allowing only $\varphi$-worlds that also satisfy $\psi$ to have the same rank as the most preferred $\varphi$-world. This means that the normality default has absolutely nothing to say about the ranks of the less preferred $\varphi$-worlds. Furthermore in system $Z^+$ maximal ignorance is assumed and the $\kappa^+$ ranking assigns the lowest possible rank permitted by the admissibility constraints. Consequently the admissibility constraints become the *only* constraints that are imposed by the normality defaults. This leads $Z^+$ to fail to sanction inheritance of properties across exceptional subclasses; a desirable behavior that is present in competing default systems like conditional entailment [Geffner, 1989], system $Z^*$ [Goldszmidt et al., 1990, Goldszmidt, 1992] (maximum entropy) and $CO^*$ [Boutilier, 1992].

An example will illustrate the problem. Consider the defaults database containing the defaults "typically birds fly", $b \to f$ and "birds typically have legs", $b \to l$. The $\kappa^+$ ranking is given by

$$\kappa^+(\omega) = \begin{cases} 1 & \text{if } \omega \models b \wedge (\neg f \vee \neg l) \\ 0 & \text{otherwise.} \end{cases}$$

If we are to discover a bird that does not fly we would be unable to conclude that it has legs because $\kappa^+$ assigns the same ranks to $\omega = b\overline{f}l$ and $\nu = b\overline{fl}$. Given the above defaults, we would prefer to believe in $\omega$ rather than $\nu$ since

1. $\omega$ verifies $b \to l$,
2. $\nu$ falsifies $b \to l$ and
3. they agree on $b \to f$.

This reflects the assumption of maximal independence where the additional information about flying is assumed to be irrelevant (unless it is the cause of disagreement with respect to some other defaults in the database). This leads us to strengthen the $\kappa^+$ interpretation of normality defaults.

## 3  Ceteris Paribum Admissibility

Consider the normality default $b \to l$ and the worlds $\omega = b\overline{f}l$ and $\nu = b\overline{fl}$. Since $\omega$ and $\nu$ differ only in their assignment to $l$, we will like to be able to infer from $b \to l$ that belief in $\omega$ is preferred to belief in $\nu$ where possible, that is $\kappa(\nu) > \kappa(\omega)$. However this inference may not always be consistent with the rest of the defaults database. For example if non-flying animals typically do not have legs, $\neg f \to \neg l$ then we may also want to infer $\kappa(\omega) > \kappa(\nu)$. However, if two worlds, $\omega$ and $\nu$, agree on all the defaults, except for those in a set $S$ which are all verified by $\omega$ but falsified by $\nu$, then we can safely prefer belief in $\omega$ over belief in $\nu$. This motivates the definition of *cp-conditions*: binary relations between worlds that agree on all the normality defaults outside of a set whose defaults are verified by one world but falsified by the other.

**Definition 3 (CP Conditions)** *Let $\Delta$ be a set of normality defaults and let $\delta$ be the maximum degree of*

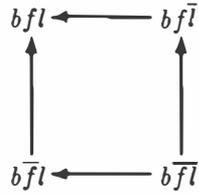

Figure 1: The cp-conditions of $\{b \to f, b \to l\}$.

*the defaults in $\Delta' \subseteq \Delta$. If worlds $\omega$ and $\nu$ are such that $\omega$ verifies all the defaults in $\Delta'$ and $\nu$ falsifies all the defaults in $\Delta'$ and $\omega$ and $\nu$ agree on all the other defaults in $\Delta \setminus \Delta'$ then $\omega >_\delta \nu$ is a **cp-condition** of $\Delta$.*

The first theorem tells us that cp-conditions are transitive.

**Theorem 1 (Transitivity of cp-conditions)** *If $\omega >_{\delta'} \omega'$ and $\omega' >_{\delta''} \omega''$, then $\omega >_\delta \omega''$ where $\delta = \max(\delta', \delta'')$.*

As an illustration cp-conditions, let us consider the defaults database, $\{d_1 = b \to f, d_2 = b \to l\}$. Worlds $bfl$ and $b\overline{f}l$ agree on $d_2$ and disagree on $d_1$ while $b\overline{f}l$ and $b\overline{f}\overline{l}$ agree on $d_1$ and disagree on $d_2$. Therefore we have cp-conditions $bfl < b\overline{f}l < b\overline{f}\overline{l}$ (with default degrees of 0). Similarly $bfl < bf\overline{l} < b\overline{f}\overline{l}$ are cp-conditions. Figure 1 shows these cp-conditions, where $\omega \to \nu$ represents $\omega > \nu$. (Since cp-conditions are transitive, $bfl < b\overline{f}\overline{l}$ is also a cp-condition.)

**Definition 4 (CP Admissibility)** *Let $\Delta$ be a set of normality defaults. A belief ranking $\kappa$ is **cp-admissible** with respect to $\Delta$ if and only if $\kappa$ is admissible and*
$$\kappa(\omega) > \kappa(\nu) + \delta$$
*for all cp-conditions $\omega >_\delta \nu$ of $\Delta$. $\Delta$ is **cp-consistent** if there exists a belief ranking that is cp-admissible with respect to $\Delta$.*

The requirement for $\kappa$ to be admissible (in the $Z^+$ sense) is not a redundant condition in the sense that not all belief rankings that satisfy the cp-conditions are admissible. Consider the set $\Delta = \{b \to f, p \to \neg f, p \to b\}$. The defaults $p \to \neg f$ and $p \to b$ are satisfied by the worlds $b\overline{p}f$ and $b\overline{p}\overline{f}$. As for $b \to f$, it is verified by $b\overline{p}f$ and falsified by $b\overline{p}\overline{f}$. Therefore $b\overline{p}f < b\overline{p}\overline{f}$ is a cp-condition of $\Delta$. Also both worlds $\overline{b}pf$ and $bpf$ agree on $p \to b$ (falsified) and $b \to f$ (satisfied) while disagreeing on $p \to f$. So $\overline{b}p\overline{f} < \overline{b}pf$ is also a cp-condition of $\Delta$. Thus the cp-conditions of $\Delta$ are $b\overline{p}f < b\overline{p}\overline{f}$ and $\overline{b}p\overline{f} < \overline{b}pf$; both due to the normality default $p \to \neg f$. The belief ranking
$$\kappa(\omega) = \begin{cases} 1 & \text{if } \omega = b\overline{p}\overline{f} \text{ or } \omega = \overline{b}pf \\ 0 & \text{otherwise} \end{cases}$$
satisfies both cp-conditions but is not admissible.

Despite the additional constraints on admissibility, cp-consistency turns out to be no different from the original notion of consistency. This next theorem tells us precisely that.

**Theorem 2 (Consistency Equivalence)** *A set of normality defaults $\Delta$ is consistent if and only if it is cp-consistent.*

In [Goldszmidt, 1992, p. 25], a procedure for testing the consistency of a database of normality defaults was presented. The equivalence of cp-consistency and consistency in $Z^+$ implies that the same procedure may be used to check for cp-consistency. The complexity of the procedure is $O(|\Delta|^2)$ satisfiability tests on the material counterparts[1] of the normality defaults in $\Delta$ [Goldszmidt and Pearl, 1991]. Although the propositional satisfiability problem is NP-complete, Horn clauses are known to admit tractable algorithms [Dowling and Gallier, 1984]. Therefore the procedure is tractable for defaults databases that have Horn material counterparts.

Even with the stronger notion of cp-admissibility, a set of normality defaults will typically still admit a large set of belief rankings. To be able to make stronger conclusions, we select a distinguished ranking, in this case the $\bar{\kappa}$ belief ranking in which we retain the assumption of maximal ignorance (as in system $Z^+$) and assign every world the lowest possible rank. First we observe that cp-admissibility is closed under minimization.

**Theorem 3 (Minimization)** *Let $\Delta$ be a set of normality defaults and let $\kappa_1$ and $\kappa_2$ be two belief rankings. If $\kappa_1$ and $\kappa_2$ are cp-admissible then $\kappa = \min(\kappa_1, \kappa_2)$ is cp-admissible.*

**Definition 5 (Minimal Rankings)** *Let $K$ be a set of belief rankings. $\kappa$ is a **minimal** ranking in $K$ if for all other rankings $\kappa' \in K$*
$$\kappa'(\omega) > \kappa(\omega)$$
*for some world $\omega$.*

Theorem 3 implies that a set $K$ of cp-admissible belief rankings of $\Delta$ has a unique minimal belief ranking given by
$$\bar{\kappa}(\omega) = \min\{\kappa(\omega) \mid \kappa \in K\}.$$
Thus the $\bar{\kappa}$ ranking of $\Delta$ is the belief ranking that assigns the lowest rank to every world among the cp-admissible rankings of $\Delta$.

**Definition 6 (The $\bar{\kappa}$ Ranking)** *Let $\Delta$ be a consistent set of normality defaults. The $\bar{\kappa}$ belief ranking is a cp-admissible ranking that is minimal in the sense that for all worlds $\omega$*
$$\bar{\kappa}(\omega) \leq \kappa(\omega)$$
*for all $\kappa$ that is cp-admissible with respect to $\Delta$.*

---

[1] The material counterpart of $\varphi \xrightarrow{\delta} \psi$ is the wff $\varphi \supset \psi$.

Table 1: Comparison between $\kappa^+$ and $\bar{\kappa}$ rankings.

| Ranks | $\kappa^+$ | $\bar{\kappa}$ |
|---|---|---|
| 0 | $\bar{b}\bar{p}\bar{f}, \bar{b}\bar{p}f, b\bar{p}f$ | $\bar{b}\bar{p}\bar{f}, \bar{b}\bar{p}f, b\bar{p}f$ |
| 1 | $b\bar{p}\bar{f}, bp\bar{f}$ | $b\bar{p}\bar{f}, bp\bar{f}$ |
| 2 | $bpf, \bar{b}pf, \bar{b}p\bar{f}$ | $bpf, \bar{b}pf$ |
| 3 |  | $\bar{b}p\bar{f}$ |

Table 2: $\bar{\kappa}$ ranking of $\Delta \cup \{b \rightarrow w\}$.

| Ranks | Worlds |
|---|---|
| 0 | $\bar{b}\bar{p}\bar{f}w, \bar{b}\bar{p}fw, b\bar{p}fw, \bar{b}\bar{p}\bar{f}\bar{w}, \bar{b}\bar{p}f\bar{w}$ |
| 1 | $b\bar{p}\bar{f}w, bp\bar{f}w, b\bar{p}\bar{f}\bar{w}$ |
| 2 | $b\bar{p}\bar{f}\bar{w}, bp\bar{f}\bar{w}, bpfw, \bar{b}p\bar{f}w, \bar{b}p\bar{f}\bar{w}$ |
| 3 | $\bar{b}pfw, \bar{b}pf\bar{w}, bpf\bar{w}$ |

As in [Goldszmidt, 1992], each belief ranking $\bar{\kappa}$ defines a consequence relation $\models_{\bar{\kappa}}$ where $\phi \models_{\bar{\kappa}} \sigma$ if and only if $\kappa(\sigma \wedge \phi) < \kappa(\neg\sigma \wedge \phi)$ or if $\kappa(\phi) = \infty$. The basic idea is to verify that $\sigma$ holds in all the minimally ranked models of $\phi$.

Let us consider our database $\Delta$ of normality defaults $\{b \rightarrow f, p \rightarrow \neg f, p \rightarrow b\}$ concerning *birds, penguins* and *flying*. As discussed above, the cp-conditions of $\Delta$ are $b\bar{p}f < b\bar{p}\bar{f}$ and $\bar{b}p\bar{f} < \bar{b}pf$; both due to the normality default $p \rightarrow \neg f$. The $\kappa^+$ and $\bar{\kappa}$ rankings are shown in table 1. We see that the world $\bar{b}pf$ is forced to a higher rank because of the cp-condition $\bar{b}p\bar{f} < \bar{b}pf$. Therefore one conclusion that we can draw from the $\bar{\kappa}$ ranking is that non-bird penguins do not fly. This conclusion escapes the $\kappa^+$ belief ranking as $\bar{b}pf$ and $\bar{b}p\bar{f}$ are assigned the same $\kappa^+$ rank.

Now if we add the default "birds typically have wings", $b \rightarrow w$ to $\Delta$, we will obtain the belief ranking shown in table 2. The cp-conditions are

$$\begin{aligned}
b\bar{p}f\bar{w} &> b\bar{p}fw \\
b\bar{p}\bar{f}\bar{w} &> b\bar{p}\bar{f}w \\
bp\bar{f}\bar{w} &> bp\bar{f}w \\
bpf\bar{w} &> bpfw \\
b\bar{p}\bar{f}\bar{w} &> b\bar{p}f\bar{w} \\
b\bar{p}\bar{f}w &> b\bar{p}fw \\
\bar{b}pf\bar{w}, \bar{b}pfw &> \bar{b}p\bar{f}\bar{w}, \bar{b}p\bar{f}w.
\end{aligned}$$

The first four cp-conditions are due to the default $b \rightarrow w$, the next two are due to $b \rightarrow f$ while the last set is due to $p \rightarrow \neg f$. We note that the minimally ranked $p$-world is $bp\bar{f}w$. Therefore, if we are to discover a penguin, we would conclude from the $\bar{\kappa}$ ranking that it is winged non-flying bird. When compared to the $\kappa^+$ ranking, we see that the default $b \rightarrow w$ imposes the cp-condition $bpf\bar{w} > bpfw$, thereby admitting the additional conclusion. Thus despite being an exceptional bird with respect to flying, penguins are still allowed to inherit the other properties associated with its birdness.

It is unclear at this point in time, if the computation of the *kappa* ranking is computationally more complex that the computation of the $\kappa^+$ ranking. We do not have a procedure for computing the $\bar{\kappa}$ ranking from a defaults database.

## 4  Related Work

Table 3: Status of defaults with respect to $\bar{b}pfw$ and $\bar{b}pf\bar{w}$.

| Default | $Z^+$ priority | $\bar{b}pfw$ | $\bar{b}pf\bar{w}$ |
|---|---|---|---|
| $b \rightarrow f$ | 1 | N | N |
| $p \rightarrow \neg f$ | 2 | F | V |
| $p \rightarrow b$ | 2 | F | F |
| $\xrightarrow{1} w$ | 2 | V | F |

$V$ = verified, $F$ = falsified and $N$ = neither.

Boutilier [Boutilier, 1992] proposed a system that sanctions inheritance across exceptional subclasses. It combines the ordering of system $Z$ (the flat version of system $Z^+$ where all the normality defaults have degree 1), with Brewka's [Brewka, 1989] notion of preferred subtheories. While system $Z^+$ assigns the same rank to any two worlds that falsify a default with priority $z$ and no default with higher priority, Boutilier's proposal will make further comparisons. Considering only the defaults with priority $z$, if the set of defaults that are violated by one world is properly contained in the set of defaults violated by the other then the former world is preferred to the latter. This criterion turns out to be inadequate when the set of defaults is not flat. Consider the penguin database, $\Delta = \{b \rightarrow f, p \rightarrow \neg f, p \rightarrow b\}$. Suppose you add the default $\xrightarrow{1} w$, "most of the creatures under consideration have wings (with strength 1)". Table 3 shows the status of the defaults in the database with respect to two worlds $\omega = \bar{b}pfw$ and $\nu = \bar{b}pf\bar{w}$. In Boutilier's system, neither world is preferred to the other as each falsifies a default that the other verifies. Thus although the only difference between $\omega$ and $\nu$ is that $\omega$ falsifies $p \rightarrow \neg f$ while $\nu$ falsifies $\xrightarrow{1} w$, a default of greater degree, Boutilier's system is unable to distinguish the two worlds. In contrast, the cp-conditions $\omega >_1 \bar{b}p\bar{f}w$ and $\nu >_1 \bar{b}p\bar{f}w$ forces the $\bar{\kappa}$ rank of $\nu$ to be greater than the $\bar{\kappa}$ rank of $\omega$. This behavior is consistent with the conclusions (in both our system and Boutilier's system) in the simpler database $\{p \rightarrow \neg f, \xrightarrow{1} w\}$.

Geffner's conditional entailment [Geffner, 1989] induces a partial order on interpretations from the priorities of the normality defaults. Among the defaults that are falsified by only one of the two worlds $\omega$ and $\nu$, if all those defaults that are of highest priorities are falsified by the same world, say $\omega$, then the world $\omega$ is preferred to the world $\nu$. Formally, if $F[\omega]$ and $F[\nu]$ are the defaults that are falsified by $\omega$ and $\nu$ respectively, then $\omega$ is preferred to $\nu$ if and only if $F[\omega] \neq F[\nu]$ and for every default in $F[\omega] \setminus F[\nu]$ there exists a default in $F[\nu] \setminus F[\omega]$ that has a higher priority. Thus if the world $\omega$ falsifies a proper subset of the defaults falsified by $\nu$ then $\omega$ is preferred to $\nu$. In our system, we impose a cp-condition quantified by the maximum degree of the defaults in $F[\nu] \setminus F[\omega]$ when $F[\omega]$ is a proper subset of $F[\nu]$ and the worlds $\omega$ and $\nu$ satisfy the additional conditions that they agree on all the other defaults.

Another difference is that unlike the numeric $Z^+$, priorities in conditional entailment is a binary relation and in general gives rise to a partial order among defaults. If a set of defaults does not tolerate default $d$ then at least one default in the set has a lower priority than $d$. This embodies the idea, that a default should have a higher priority than a composite argument to the contrary, which follows naturally from the interpretation of a default $\varphi \to \psi$ as "if $\varphi$ is all we know then we are authorized to assert $\psi$, regardless all the other normality defaults in the database". As a result of this partial order among defaults it becomes non trivial to extend conditional entailment to take into account the degrees of quantified normality defaults.

Pearl, motivated by the connection [Jaynes, 1979, Tribus, 1969] between maximizing entropy and minimizing dependencies, proposed [Pearl, 1988, p. 491] that the maximum entropy principle could be used to extract implicit independencies in default reasoning. Taking such an approach, Goldszmidt [Goldszmidt, 1992, Goldszmidt et al., 1990] proposed a system that ranks a world according to the weighted sum of the defaults falsified by the world. System $Z^+$ ranks a world according to the maximum priority default that is falsified by the world. By incorporating the cp-conditions, we have introduced some form of summation of the degrees of the falsified defaults. The cp-conditions are however only between worlds that agree on the other defaults in the database.

In [Selman and Kautz, 1988] Selman and Kautz introduced systems of Propositional Model Preference Defaults where defaults of the form $\alpha \to q$, where $\alpha$ is a wff and $q$ is a literal, are considered. Each default $\alpha \to q$ induces a "ceteris paribum" preference between worlds that agree on all the atomic propositions with the possible exception of the proposition occurring in $q$. This extremely local (local to a single default) view prevents the system handling specificity properly. In contrast, in the $\bar{\kappa}$ interpretation, a default induces a cp-condition between worlds only when they agree on all the other defaults in the database.

## 5 Conclusion

In this paper we have proposed an extension to System $Z^+$ that rectifies its main shortcoming by introducing cp-conditions between worlds that agree on all the other defaults in the database. We show that the additional constraints do not change the notion of consistency of a defaults database in system $Z^+$. This means that the semi-tractable algorithm for determining consistency may also be used to check for cp-consistency. It turns out that the main difference between our system and many of the other default reasoning systems is that we take a more "global" view, placing constraints between worlds only when they agree on all the other defaults in the database. The advantages and disadvantages of this global view with respect to the other default reasoning systems remains to be fully evaluated.

## Acknowledgements

We would like to thank three anonymous reviewers for their constructive comments and suggestions. The first author is supported in part by a scholarship from the National Computer Board, Singapore. The research was partially supported by Air Force grant #AFOSR 90 0136, NSF grant #IRI-9200918, and Northrop-Rockwell Micro grant #93-124.

## A Proofs

**Theorem 1 (Transitivity of cp-conditions)** If $\omega \succ_{\delta'} \omega'$ and $\omega' \succ_{\delta''} \omega''$, then $\omega \succ_{\delta} \omega''$ where $\delta = \max(\delta', \delta'')$.

**Proof:** Let $\Delta$ be the defaults database, $S'$ be the set of defaults that $\omega$ and $\omega'$ disagree on and $S''$ be the that $\omega'$ and $\omega''$ disagree on. Consider the set $S = S' \cup S''$ and a default $d$. First we note that $S'$ and $S''$ are disjoint. If $d \in S'$, then by definition it is falsified by $\omega$ and verified by $\omega'$. Verification of $d$ by $\omega'$ implies verification by $\omega''$ since the worlds $\omega'$ and $\omega''$ agree outside of $S''$. If $d \in S''$, then by definition it is falsified by $\omega'$ and verified by $\omega''$. In this case, falsification by $\omega'$ implies falsification by $\omega$ since the worlds $\omega$ and $\omega'$ agree outside of $S'$. Agreement by $\omega$ and $\omega'$ on $\Delta \setminus S'$ and by $\omega'$ and $\omega''$ on $\Delta \setminus S''$ also imply that $\omega$ and $\omega''$ agree on $\Delta \setminus S$. Therefore $\omega \succ_\delta \omega''$ where $\delta$ is the maximum degree in $S$ i.e. $\max(\delta', \delta'')$. □

**Lemma 1 (Minimization)** Let $\Delta$ be a set of normality defaults and let $\kappa_1$ and $\kappa_2$ be two belief rankings. If $\kappa(\omega) = \min(\kappa_1(\omega), \kappa_2(\omega))$ then $\kappa(\phi) = \min(\kappa_1(\phi), \kappa_2(\phi))$.

**Proof:**

$$\begin{aligned} \kappa(\phi) &= \min\{\kappa(\omega) \mid \omega \models \phi\} \\ &= \min\{\min(\kappa_1(\omega), \kappa_2(\omega)) \mid \omega \models \phi\} \\ &= \min(\min\{\kappa_1(\omega) \mid \omega \models \phi\}, \end{aligned}$$

$$\min\{\kappa_2(\omega) \mid \omega \models \phi\})$$
$$= \min(\kappa_1(\phi), \kappa_2(\phi)).$$

**Theorem 3 (Minimization)** Let $\Delta$ be a set of normality defaults and let $\kappa_1$ and $\kappa_2$ be two belief rankings. If $\kappa_1$ and $\kappa_2$ are cp-admissible then $\kappa = \min(\kappa_1, \kappa_2)$ is cp-admissible.

**Proof:** First we will show that $\kappa$ is admissible.

$$\begin{aligned}\kappa(\varphi \wedge \neg\psi) &= \min(\kappa_1(\varphi \wedge \neg\psi), \kappa_2(\varphi \wedge \neg\psi))\\ & \qquad\qquad\text{(by lemma 1)}\\ &\geq \min(\kappa_1(\varphi \wedge \psi) + \delta, \kappa_2(\varphi \wedge \psi) + \delta)\\ & \qquad\qquad\text{(by admissibility of }\kappa_i)\\ &= \min(\kappa_1(\varphi \wedge \psi), \kappa_2(\varphi \wedge \psi)) + \delta\\ &= \kappa(\varphi \wedge \psi) + \delta \qquad\text{(by lemma 1).}\end{aligned}$$

Next we show that all cp-conditions are satisfied. Let $\omega \succ_\delta \nu$.

$$\begin{aligned}\kappa(\omega) &= \min(\kappa_1(\omega), \kappa_2(\omega))\\ &\geq \min(\kappa_1(\nu) + \delta, \kappa_2(\nu) + \delta)\\ & \qquad\text{(by cp-admissibility of }\kappa_i)\\ &= \min(\kappa_1(\nu), \kappa_2(\nu)) + \delta\\ &= \kappa(\nu) + \delta.\end{aligned}$$

Therefore cp-admissibility is closed under minimization. □

**Lemma 2 (Goldszmidt [Goldszmidt, 1992])** *Let $\Delta$ be a set of normality defaults. If $\Delta$ is consistent then the defaults in $\Delta$ can be partitioned into $\Delta_0, \ldots, \Delta_m$ such that all normality defaults $d \in \Delta_i$ is tolerated by $\Delta \setminus \bigcup_{j=0}^{i-1} \Delta_j$.[2] Furthermore there is a sequence of sets of worlds $\Omega_i \subset \Omega$ such that for all $\omega \in \Omega_i$, $\omega$ verifies all the normality defaults in $\Delta_i$ and $\omega$ satisfies $\Delta \setminus \bigcup_{j=0}^{i-1} \Delta_j$.*

**Proof:** This lemma follows directly from the procedure for testing consistency [Goldszmidt, 1992, p. 25]. (See also corollary 2.5 there.) □

**Lemma 3 (Consistency of CP-Conditions)** *The cp-conditions of any set $\Delta$ do not form a $<$-cycle; that is we cannot find a sequence of worlds such that $\omega_0 <_{\delta_0} \ldots <_{\delta_{n-1}} \omega_n <_{\delta_n} \omega_{n+1} = \omega_0$ for any $n$.*

**Proof: (By contradiction)** Suppose we can find such a sequence. Then for all $i = 0, \ldots, n$, $\omega_i <_{\delta_i} \omega_{i+1}$ implies that there exists a normality default $d_i$ such that $\omega_i$ verifies $d_i$ and $\omega_{i+1}$ falsifies $d_i$. In addition $\omega_i$ and $\omega_{i+1}$ agrees on all the other defaults. Therefore $\omega_i$ verifies $d_i$ and falsifies $d_j$ for all $j < i$. In particular, $\omega_{n+1}$ falsifies $d_0$ but $\omega_{n+1} = \omega_0$ implies that $\omega_{n+1}$ also verifies $d_0$. This is a contradiction. □

**Theorem 2 (Consistency Equivalence)** A set of normality defaults $\Delta$ is consistent if and only if it is cp-consistent.

---

[2]If $i < 1$ then $\bigcup_{j=0}^{i-1} S_j = \emptyset$.

**Proof:** Since a cp-admissible ranking is also admissible with respect to $\Delta$, cp-consistency of $\Delta$ implies consistency of $\Delta$. Next we assume that $\Delta$ is consistent and construct a belief ranking $\kappa$ that is cp-admissible. For $i = 0, \ldots, m$, (by lemma 2) we construct $\Delta_i$ and $\Omega_i$ such that all normality defaults $d \in \Delta_i$ is tolerated by $\Delta \setminus \bigcup_{j=0}^{i-1} \Delta_j$ and for all $\omega \in \Omega_i$, $\omega$ verifies all the normality defaults in $\Delta_i$ and $\omega$ satisfies $\Delta \setminus \bigcup_{j=0}^{i-1} \Delta_j$. We also define $\Omega_{m+1}$ to be $\Omega \setminus \bigcup_{j=0}^{m} \Omega_j$. Now we partition each of the $\Omega_i$'s. Defining $\omega$ to be $<$-minimal in $S$ if for all $\nu \in S, \nu \not< \omega$, we construct $\Omega_{i,j}$ as follows.

$$\Omega_{i,j} = \{\omega \mid \omega \text{ is } <\text{-minimal in } \Omega_i \setminus \bigcup_{l=0}^{j-1} \Omega_{i,l}\}.$$

This construction is possible as there are no $<$-cycles by lemma 3. Let $\delta_i^*$ denote the largest degree of defaults in $\Delta_i$. We define the belief ranking $\kappa(\omega)$ for each $\omega$ in $\Omega_{i,j}$ as

$$\kappa(\omega) = j \times \delta_i^* + \max_{\nu \in \Omega_{i-1}} \kappa(\nu) + \delta_{i-1}^*$$

with the second term being 0 when $i = 0$.

We will show that the belief ranking $\kappa$ is cp-admissible. First we consider worlds $\omega \in \Omega_j$ and $\nu \in \Omega_i$ where $i < j$.

$$\begin{aligned}\kappa(\omega) &\geq \max_{\nu \in \Omega_{j-1}} \kappa(\nu) + \delta_{j-1}^*\\ &\geq \max_{\nu \in \Omega_{j-2}} \kappa(\nu) + \delta_{j-2}^* + \delta_{j-1}^*\\ &\;\;\vdots\\ &\geq \max_{\nu \in \Omega_i} \kappa(\nu) + \sum_{l=i}^{j-1} \delta_l^*\\ &\geq \kappa(\nu) + \sum_{l=i}^{j-1} \delta_l^*.\end{aligned}$$

Therefore if $i < j$

$$\kappa(\omega) \geq \kappa(\nu) + \delta_i^* \qquad (1)$$

for all $\omega \in \Omega_j$ and $\nu \in \Omega_i$. Given a normality default $d = \varphi \xrightarrow{\delta} \psi \in \Delta_i$, we have $\kappa(\varphi \wedge \psi) \leq \kappa(\nu)$ for some $\nu \in \Omega_i$ because $\nu \in \Omega_i$ implies that $\nu \models \varphi \wedge \psi$. If $\omega \models \varphi \wedge \neg\psi$ then $\omega \in \Omega_j$ for some $j > i$. Then for all $\nu \in \Omega_i$ (by equation 1),

$$\begin{aligned}\kappa(\omega) &\geq \kappa(\nu) + \delta_i^*\\ &\geq \kappa(\nu) + \delta\\ &\geq \kappa(\varphi \wedge \psi) + \delta.\end{aligned}$$

Therefore $\kappa$ is admissible.

Let us consider the cp-condition $\omega >_\delta \nu$ with $\nu \in \Omega_i$ and $\omega \in \Omega_j$. $\omega$ and $\nu$ agree on all defaults except for some $d \in \Delta_l$ which $\omega$ falsifies and $\nu$ verifies. Since $\omega$ falsifies $d$ we know that $l < j$. The agreement of $\omega$ and $\nu$ on the defaults in $\Delta_m$ for $m > l$ implies that $i \leq j$. If $i < j$ then $\kappa(\omega) \geq \kappa(\nu) + \delta$ (by equation 1). If

Exceptional Subclasses in Qualitative Probability 559

$i = j$ then $\omega >_\delta \nu$ implies that $\omega \in \Omega_{i,x}$ and $\nu \in \Omega_{i,y}$ for some $x > y$. Then $\kappa(\omega) - \kappa(\nu) = (x - y)\delta_i^* \geq \delta$. Therefore $\kappa(\omega) \geq \kappa(\nu) + \delta$. This completes the proof. □